\documentclass[runningheads]{llncs}
\usepackage[T1]{fontenc}
\usepackage{graphicx}
\usepackage{hyperref}
\usepackage{amsmath,amssymb,mathtools} 
\usepackage[misc]{ifsym}
\usepackage{csquotes}
\usepackage{multirow,booktabs}
\usepackage{tikz, ifthen}
\usepackage[normalem]{ulem}

\newcommand{\abc}[1]{(\textsf{\bfseries #1})}
\newcommand{\myparagraph}[1]{\subsubsection{#1}}

\newboolean{anonymize}
\setboolean{anonymize}{false}

\definecolor{maroon}{rgb}{0.5, 0.0, 0.0}
\definecolor{green}{rgb}{0, 0.5, 0}

\begin{document}
\title{Fast and Adaptive Questionnaires for \texorpdfstring{\\}{}Voting Advice Applications}
%
%

\ifthenelse{\boolean{anonymize}}
    {
        \author{Anonymous Authors}
        \authorrunning{Anonymous Authors}
        \institute{Anonymous Institute}
        \def\pypipackage{at [PyPi link]}
        \def\githubpaper{at [GitHub link]}
        \def\githubpackage{at [GitHub link to be added after review]}
    }{
        \author{Fynn Bachmann~\href{mailto:fynn.bachmann@uzh.ch}{\Letter}\and
        Cristina Sarasua\and
        Abraham Bernstein}
        \authorrunning{Bachmann et al.}
        %
        \institute{University of Zurich, Switzerland\\
        Department of Informatics\\
        \email{\{fbachmann,sarasua,bernstein\}@ifi.uzh.ch}}
    }

\maketitle              
\begin{abstract}

The effectiveness of Voting Advice Applications (VAA) is often compromised by the length of their questionnaires. To address user fatigue and incomplete responses, some applications (such as the Swiss Smartvote) offer a condensed version of their questionnaire. However, these condensed versions can not ensure the accuracy of recommended parties or candidates, which we show to remain below 40\%. To tackle these limitations, this work introduces an adaptive questionnaire approach that selects subsequent questions based on users' previous answers, aiming to enhance recommendation accuracy while reducing the number of questions posed to the voters. 
Our method uses an encoder and decoder module to predict missing values at any completion stage, leveraging a two-dimensional latent space reflective of political science's traditional methods for visualizing political orientations. Additionally, a selector module is proposed to determine the most informative subsequent question based on the voter's current position in the latent space and the remaining unanswered questions. 
We validated our approach using the Smartvote dataset from the Swiss Federal elections in 2019, testing various spatial models and selection methods to optimize the system's predictive accuracy. Our findings indicate that employing the IDEAL model both as encoder and decoder, combined with a \textsf{PosteriorRMSE} method for question selection, significantly improves the accuracy of recommendations, achieving 74\% accuracy after asking the same number of questions as in the condensed version. 
%
\end{abstract}
\section{Introduction}

Voting Advice Applications (VAA) have emerged as a crucial tool for voters in multi-party democracies~\cite{garzia_voting_2019}. By comparing answers to a questionnaire, VAAs offer guidance on which candidates or parties best align with voters' political preferences -- considerably affecting voter turnout and election results~\cite{pianzola_impact_2019}. In Switzerland, for example, the VAA Smartvote\footnote{\url{https://smartvote.ch}} increases the probability of voters casting their vote by 12\%~\cite{garzia_voting_2017}.
%
With such an impact it becomes even more important that the recommendations remain accurate, even if voters skip questions or exit the questionnaire early due to user fatigue~\cite{orsini2022machine}. To this end, Smartvote offers a \enquote{rapid version} of its questionnaire, where 31 out of 75 questions were manually selected to capture at least the most relevant answers for accurate recommendations. However, we show that this rapid version might only yield an accuracy of 40\% compared to what would have been recommended if a voter had given all answers. 

Instead, we propose to use an adaptive questionnaire that selects the next question based on previous answers of a voter. To achieve this, we use the combination of (i) an encoder that embeds users into a latent space based on few answers, (ii) a decoder that predicts future answers based on the location in the latent space, and (iii) a selector that chooses the next question to be asked based on the expected information gain. Additionally, we suggest including the predictions to unanswered questions in the matching formula of the recommendations. We hypothesize that by doing so the recommendations will be more accurate than by entirely leaving out unanswered questions. This is a novel approach that is directly applicable even to static questionnaires.

Related work has explored adaptive questionnaires in similar but not identical settings. One study on adaptive testing in political surveys identified the absence of training data as a significant hurdle~\cite{montgomery_computerized_2013}. In the context of VAA, however, training data arise naturally through the answers of the parties or candidates who fill out the questionnaire beforehand.  In another, very recently published study, adaptive questionnaires outperformed static questionnaires for party recommendations in VAAs~\cite{sigfrid_irt_2024}. However, their method recommends the closest party purely based on the position in the latent space. This approach fails to accurately recommend candidates from a large set of choices due to the loss of information in the compression, only giving an accuracy of 16\% even if all questions were answered. For comparison, we cover a variant of their method under \textsf{PosteriorVariance} in our analysis.

Our study consists of two experiments, both utilizing the dataset from the Smartvote VAA of the Swiss Federal election in 2019~\cite{smartvote2019data}. In the first experiment, we analyze a wide range of dimensionality reduction algorithms for their ability to encode and decode sparse data (e.g., principal component analysis, item-response theory~\cite{clinton_statistical_2004}, variational autoencoders~\cite{mccoy_variational_2018}). 
In the second experiment, we then compare different question selection methods to optimize the questionnaire (e.g. uncertainty, k-nearest neighbors, posterior variance).
The overall analysis shows that the best combination of encoder, decoder, and selector can drastically improve the recommendation accuracy: specifically, it can (i) produce the same degree of accuracy as the current rapid version with only 9 rather than 31 questions or (ii) it achieves an accuracy over 74\% (in contrast to 40\%) when asking the same number of questions as the rapid version. 
Furthermore, we find that considering the spatial model's predictions to unanswered questions for candidate recommendations can improve the accuracy by around 7\% -- regardless of the order of the questionnaire.
We conclude that our methods represent a viable enhancement to VAAs, offering more accurate recommendations with fewer questions while maintaining a transparent process for voters and researchers.


\section{Methods}
\label{sec:methods}

Our study relies on the Smartvote data from the Swiss Federal elections in 2019, spatial models (or embeddings) to understand a user's preference, a set of selection methods to choose the next most desirable question, and some standard evaluation metrics -- all of which this section visits in turn.   

\subsection{Data}
\label{sec:methods_data}

We consider the VAA dataset of the Swiss federal election of 2019 obtained from Smartvote. The data contain the answers of 4663 Swiss political candidates to 75 representative questions about the political agenda. These answers are given on a Likert scale, where questions offer 4 to 7 ordinal options (i.e., from \emph{fully disagree} to \emph{fully agree}). In the data matrix $R$, Likert scale values are provided as numbers between 0 and 1, where 0 corresponds to \emph{fully disagree}, 1 to \emph{fully agree}, and the remaining answers are evenly distributed across the interval. To make our study also applicable for binary settings outside of VAA (e.g., roll-call data~\cite{boche_new_2018} or wiki surveys such as Polis~\cite{small2021polis}), we discretized the reactions to two values such that all values $0.5\leq r\in R$ are encoded as `1' and all others as `0'. For preprocessing, we removed all candidates who have not answered all 75 questions or do not belong to one of the seven major Swiss parties:\footnote{The party affiliation does not matter for our analysis. The number of parties was merely reduced for visual purposes in the figures. A legend of party colors can be found in the appendix in Figure~\ref{fig:legend}.} BDP, CVP, FDP, GLP, GPS, SP, or SVP.\footnote{BDP and CVP have since merged to a new party called ``Die Mitte''.} The remaining number of 1912 candidates was further separated into a train and a test set, where we use 15\% of the candidates as test users ($N_{\text{test}}=290$). For each of these datasets, varying ratios $u,v \in \{0\%,10\%,...,90\%\}$ of the cells were removed from the train and test set respectively, resulting in 10 train sets $R_u$ and 10 test sets $T_v$ (likewise for the binary case). A schema of this procedure is given in the appendix in Figure~\ref{fig:schema}. To facilitate the reproducibility of the results in this study, we uploaded the anonymized data (removing candidate information) to our GitHub repository.\footnote{\url{https://github.com/fsvbach/aqvaa-paper}}

\subsection{Spatial Models}
\label{sec:methods_spatial}

In political science, spatial models are often used to visualize the ideological orientation of voters or candidates in a one or two-dimensional space -- referred to as \emph{ideal point estimation}~\cite{poole2007scaling}. From all the algorithms that could estimate such ideal points, we chose a subset that represents different approaches in item response theory~\cite{reckase2006irt} and machine learning while fulfilling two prerequisites: Firstly, each algorithm must seamlessly integrate new users into the established space without altering the model parameters defined during the training phase. Secondly, they should be able to predict answers to  questions merely based on the location in the latent space. While some of the algorithms chosen have an inherent decoder module to perform this task, we use logistic regression (LR) for the others. In these cases, one LR model is trained for each question using the coordinates in the latent space which is then able to predict probabilities (for the binary case) and can even be used as a regression model for Likert scales. 
Notably, we do not restrict the decoder modules to have linear decision boundaries  but also experiment with algorithms that have non-linear decoders (e.g. neural networks). For those algorithms, we only require that the decoder module is sufficiently smooth so that the predictions can be visually explained to potential users of the VAAs. While our selection of algorithms does not cover all approaches to the given problem, we consider our choice broad enough to assess the qualitative differences between the different fields. 
Visual examples and a legend can be found in Appendix~\ref{app:models}. 

\myparagraph{Principle Component Analysis (PCA)}
Given its fast computation, PCA is widely used to linearly project high-dimensional data into a low-dimensional space while preserving as much of the variance as possible. Recently, PCA has also been proposed in the context of ideal point estimation~\cite{potthoff_estimating_2018}. In such setting, the method is currently applied in the Smartmap from Smartvote,\footnote{
\url{https://www.smartvote.ch/de/group/527/election/23_ch_nr/smartmap}
} as well as to compute the two-dimensional representation of users' reactions in the wiki survey Polis~\cite{small2021polis}. A general drawback of PCA is, however, that it requires a complete matrix as an input. While variants of PCA exist that work with sparse train data~\cite{zou_selective_2018}, these approaches do not allow embedding a test set into the same space without modifying dimensions. We therefore mean-imputed sparse train and test sets before fitting.

An inherent method to decode the PCA embedding (i.e. to predict the  answers of candidates from latent coordinates) is using the matrix factorization (MF) of PCA. This only requires taking the outer product of the latent coordinate with the components of the projection matrix and is therefore very fast. Due to its linearity, however, it would predict values larger than one or smaller than zero for coordinates far away from the center. This approach should rather be used as a regression than to predict actual probabilities. We, therefore, additionally use LR as a decoder for this spatial model.

\myparagraph{$t$-distributed Stochastic Neighborhood Embedding ($t$-SNE)}
As a variant of multidimensional scaling, $t$-SNE is a non-linear dimensionality algorithm, which also has been applied to political datasets~\cite{van2008visualizing,bachmann_wasserstein_2023}. Based on the reactions matrix $R$, $t$-SNE computes the distance for each pair of candidates and then optimizes their location until the distances in the latent space correspond optimally to those in the original matrix. As $t$-SNE does not have a decoder module, we use LR to predict probabilities for unanswered questions from candidates' positions in the latent space. In this work, we use the \textsf{openTSNE} implementation~\cite{policar_opentsne_2019} which allows embedding a test set into an existing latent space.

\begin{figure}
\centering
\includegraphics{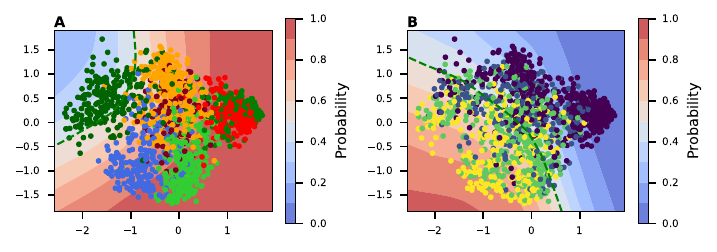}
\caption{Latent space of the VAE with a 2-layer decoder. The background color indicates model predictions. \abc{A}~Train-set candidates are  colored by their party. The non-linear decoder  predicts answers to question 27: \enquote{Should same-sex couples have the same rights as heterosexual couples in all areas?} as indicated by the background color. \abc{B}~Train-set candidates are  colored by their given answer to question 13: \enquote{Should insured persons contribute more to healthcare costs (e.g., by increasing the minimal deductible)?} The answers are 4-Likert scale from \enquote{Fully Agree} (yellow) 
to \enquote{Fully Disagree} (dark blue).}
\label{fig:vae_questions}
\end{figure}

\myparagraph{Variational Autoencoder (VAE)}

Matching our setup by design, VAEs are composed of two neural networks: an encoder network and a decoder network~\cite{kingma_auto-encoding_2022}. The encoder network computes latent coordinates in a low dimensional space, from which the decoder network aims to reconstruct the original answers. Based on the reconstruction error, the weights of the neural networks are trained until the best fit is obtained. To add variance, the VAE learns parameters for Gaussian distributions in the latent space, which can then be used to generate similar but previously unseen answer patterns to the questionnaire. For this study, we chose a 2-layer encoder network. For the decoder module, we evaluate two different networks: The first is simply a logistic regression network (linear layer with a sigmoid activation function), resulting in a linear decision boundary. The second decoder network is a 2-layer network, which should resemble an RBF-kernel able to learn non-linear prediction patterns.
In recent work, it has been proposed to use VAE with missing values~\cite{mccoy_variational_2018}. However, it remains difficult to embed sparse test data into the same latent space without imputation.  Hence, we  mean-impute all data for any $u,v>0$. In Figure~\ref{fig:vae_questions}, the VAE embedding of the train-set candidates with $u=0$ is shown with two decision boundaries of specific questions highlighted by green dashed lines. 

\myparagraph{Item-Response Theory (IRT)}

Traditionally used in educational testing to assess test-takers' abilities based on their responses,  IRT has also been widely applied to political analysis~\cite{clinton_statistical_2004,poole_spatial_1985}. In this context, abilities translate to political preferences, and test questions correspond to policy issues or votes. In training, each model identifies the user’s coordinates and item parameters that minimize the discrepancy between actual answers and those predicted by the model. This optimization involves adjusting actors' positions on the policy spectrum iteratively to fit the observed data best. Depending on the dimension of the latent space IRT models may require a lot of parameters that have to be learned, which makes them less efficient to compute. However, by design they accept sparse reaction matrices, as they will learn the parameters only based on the given interactions of candidates with questions. In this work, we evaluate two IRT ideal point estimation algorithms: IDEAL~\cite{clinton_statistical_2004,jackman2024pscl} and W-NOMINATE~\cite{poole_spatial_1985,poole2007scaling}.

\begin{figure}[t]
\centering
\includegraphics[width=\textwidth]{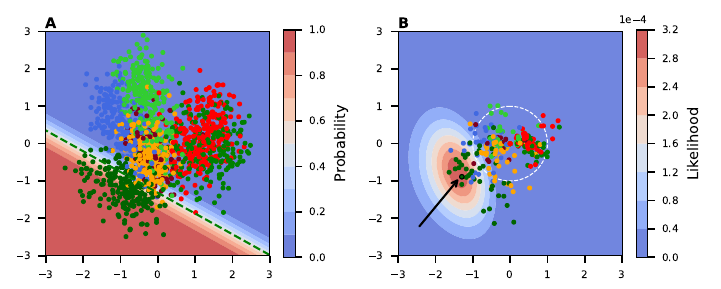}
\caption{IDEAL embedding of the binary Smartvote dataset. Dots represent candidates of their respective party. 
\abc{A}~Likelihood function $P(Y|X)$ for question 5: \enquote{Should Switzerland terminate the Schengen Agreement with the EU, in order to reintroduce more security checks directly on the border?} 
\abc{B}~Posterior distribution $P(X|Y_I)$  for test candidate 459 (highlighted by a black circle and arrow) after answering 7 random questions $y\in Y_I$. Other dots represent maximum likelihood estimators for all test candidates with equally many answers given. The dashed white circle corresponds to the standard deviation ellipse of the Gaussian prior P(X).} 
\label{fig:ideal}
\end{figure}

The first algorithm, IDEAL, is a Bayesian estimation method modeling each candidate's answer as a function of the question's position relative to their own ideal point. It incorporates a probabilistic model that accounts for uncertainty in both the candidates' positions and the question characteristics. The estimation utilizes Markov Chain Monte Carlo techniques to approximate the posterior distributions of the ideal points. The underlying formula for IDEAL reads:
\begin{equation}
    \label{eq:ideal}
    P(y_{ij} = 1) = \Phi(\beta_j X_i - \alpha_j),
\end{equation}
where $y_{ij}$ is the binary answer of candidate $i$ to question $j$, $\Phi$ is the cumulative distribution function of the normal distribution, $X_i$ represents the candidate's ideal point, and $\alpha_j, \beta_j$ are question parameters. A visual example of an IDEAL embedding is shown in Figure~\ref{fig:ideal}.

\begin{figure}[t]
\centering
\includegraphics{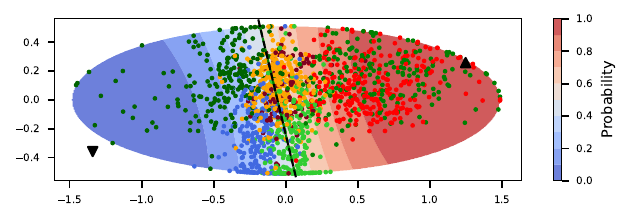}
\caption{W-NOMINATE embedding with 60\% missing data for question 6: \enquote{Should wealthy individuals contribute more to the funding of the state?} The closer candidates are to the YAY ($\blacktriangle$) and NAY ($\blacktriangledown$) locations, the more likely they vote accordingly. The 50\% decision boundary is shown as a dashed line. Candidates are colored by their party.}
\label{fig:nominate_decision}
\end{figure}

The second algorithm, W-NOMINATE (Weighted Nominal Three-Step Estimation), also positions candidates and questions within a Euclidean space but computes a utility function based on the  distance between a candidate's ideal point and a question's position. Specifically, a candidate $i$ has the utility
\begin{equation}
    u_{ijy} = \beta \text{e}^{-\frac{1}{2} d_{ijy}^2}
\end{equation}
to vote $y$ on question $j$, where $d_{ijy}$ is the $L^2$ distance between the candidate's ideal point and the position of the respective reaction, and $\beta$ is a noise parameter. The utilities of both answers are then translated to probabilities using the softmax function. Notably, W-NOMINATE restricts candidates' ideal points to lie in a unit sphere.  Since dimensions are weighted differently, this results in the elliptical shape of the W-NOMINATE embedding in Figure~\ref{fig:nominate_decision}.

\subsection{Selection Methods}
\label{sec:methods_selection}

Given a spatial model (encoder and decoder module), we defined several selection methods that decide the next question for a user with a given set of answers. Each selection method optimizes a specific quantity, which can, for example, be uncertainty, disagreement of the k-nearest neighbors, or the posterior variance (as in Figure~\ref{fig:ideal}B). In this section, we explain the objective functions of each selection method using the following notation: By $P(X)$ we denote the prior distribution for a user that has not given any answer (a normal Gaussian). $P(X|Y_I)$ is the posterior distribution for a user having answered all questions $y\in Y_I$. Furthermore, we call $P(Y_N|X)$ the likelihood of agreeing to an unanswered question $y\in Y_N$ given some location $X$ in the latent space. Finally, to quantify uncertainty for a binary probability, we take the Gini impurity 
\begin{equation}
    G(p) = 2p (1-p).
\end{equation}

\myparagraph{Fixed Orders}
As a baseline we consider selection methods that have a predefined order. First, the \textsf{DefaultOrder} corresponds to the order of the full questionnaire as it is on the Smartvote website. Second, the \textsf{RapidVersion} corresponds to the condensed questionnaire consisting of 31 manually selected questions by the Smartvote researchers. Their order was left as it is on the Smartvote website. The third fixed-order method, \textsf{FixedGini}, orders the questions by the Gini impurity of their prior  $P(Y)$, thus preferring questions with high uncertainty.

\myparagraph{k-Nearest Neighbor (kNN) methods}
A more dynamic approach is to select the next question based on the disagreement of the kNN. We evaluate two such methods, \textsf{BaseKNN} and \textsf{FullKNN}, which differ only in the way they compute the nearest neighbors: \textsf{BaseKNN} takes the kNN that would be recommended only using the given answers $Y_I$, whereas \textsf{FullKNN} predicts the answers to the remaining questions $P(Y_N|Y_I)$ first, and computes the kNN including these predictions. Both methods then select the question that maximizes the disagreement among their kNN (quantified by the Gini impurity).

\myparagraph{Probabilistic methods} 
Finally, we evaluate three probabilistic selection methods: \textsf{Uncertainty}, \textsf{PosteriorVariance}, and \textsf{PosteriorRMSE}. \textsf{Uncertainty} computes the probabilities $P(Y_N|Y_I)$ for all unanswered questions $Y_N$ given the initial answers $Y_I$. Then, it selects the question where the Gini impurity is maximal. Both \textsf{PosteriorVariance} and \textsf{PosteriorRMSE} extend this approach by additionally computing the posterior distributions. In particular, they estimate for each unanswered question $Y_k\in Y_N$ the expected posterior $P(X|Y_{I \cup {k}})$ and from that the posterior predictions $P(Y_{N\backslash {k}}|Y_{I \cup {k}})$. They then minimize the expected  posterior variance, or posterior uncertainty respectively, which results in the objective functions
\begin{equation}
    \sum_{y_k\in\{0,1\}} P(Y_k=y_k|Y_I) \cdot \text{VAR}_{X} \left[ P(X|Y_{I \cup {k}}) \right] 
\end{equation}
for \textsf{PosteriorVariance}, and
\begin{equation}
    \sum_{y_k\in\{0,1\}}  P(Y_k=y_k|Y_I) \cdot \text{VAR}_{Y_{N\backslash {k}}} \left[ P(Y_{N\backslash {k}}|Y_{I \cup {k}}) \right]
\end{equation}
for \textsf{PosteriorRMSE}. For each method, the expected information gain is computed as the difference in prior and posterior variance. A more detailed derivation of these functions and notes for their computation can be found in Appendix~\ref{app:selection}.


\subsection{Evaluation Metrics}
\label{sec:methods_metrics}
To evaluate the spatial models and their predictions, we used three different metrics. For Likert scale data, we took the root mean squared error (RMSE), which measures how far away the prediction $p$ is from an answer $y$. For the binary dataset, we measured accuracy, i.e., if the predicted probability for an answer $y\in \{0,1\}$ is $0\leq p \leq 1$, the accuracy for this user-question pair becomes 
\begin{equation}
  \alpha(p,y) = 
  \begin{cases} 
   0 & \text{if } |p-y| \geq 0.5 \\
   1 & \text{if } |p-y| < 0.5.
  \end{cases}
\end{equation}
The reason for not considering the RMSE for binary data is that we do not want to punish models for providing uncertainty, as then the best estimator would be just predicting $p=0$ or $p=1$ instead of probabilities. 
Lastly, to recommend candidates, we use the matching formula of Smartvote, which is equivalent to the $L^2$ distance of candidates' and voters' answers to the questionnaire. We then measure the candidate recommendation accuracy by set overlaps: for a recommended set of candidates $A$ and a true set $B$, we count the number that appear in both sets and divide it by their size. In this study, we chose candidate recommendations of size $32$, which is the number of representatives that citizens from the Canton of Zurich elect and that Smartvote, hence, advises them on.

\section{Results}

We conducted two experiments: In the first experiment, we evaluated seven dimensionality reduction algorithms, referred to as spatial models, for (i) their ability to generalize to test data and (ii) their robustness to sparsity. In the second experiment, we compared nine question selection methods to optimize the Smartvote questionnaire. 

\subsection{Selecting the Spatial Model}

The objective of our first experiment was to select the spatial model that could best fit the VAA dataset of Smartvote. To this end, we divided the Smartvote candidates' dataset into training and test sets, with fractions $u$ (and $v$ respectively) of the data removed as described in Section~\ref{sec:methods_data}. The training sets were encoded into a two-dimensional latent space using the different encoder modules of the spatial models (see Section~\ref{sec:methods_spatial}). We then predicted the missing values with the corresponding decoder modules. The same procedure was repeated for the (sparse) test sets using the learned model parameters. The quality of fit and imputation was assessed using two metrics: we measured accuracy for the binary dataset, and RMSE for the original dataset as explained in Section~\ref{sec:methods_metrics}. From all combinations of $u,v\in \{0\%,10\%,...,90\%\}$, we identified four relevant real-world scenarios, referred to as \textit{Train Fit}, \textit{Train Impute}, \textit{Test Fit}, and \textit{Test Impute}. All results are shown in Table \ref{tab:results}, while additional figures can be found in Appendix~\ref{app:results}.
\begin{table}
\centering
\caption{Spatial models and their averaged fit and imputation results across different sparsities. The first column describes the accuracy of the model fit of the binary train data. The second column describes the accuracy of the missing value imputation using each model with binary train data. The third and fourth columns describe the RMSE for test fit and test imputation of the original data. The baseline is computed using mean imputation. The best results are bold, second best are italicized.}
\label{tab:results}
\begin{tabular}{llllll}
\toprule
Embedding & Prediction & Train Fit & Train Imp & Test Fit  & Test Imp \\
 & & {\footnotesize \textit{Accuracy}} & {\footnotesize \textit{Accuracy}} & {\footnotesize \textit{RMSE}} & {\footnotesize \textit{RMSE}} 
 \\
\midrule
 & Mean & - & 0.663 & - & 0.380 \\
\cline{1-6}
IDEAL &  & 0.839 & \textbf{0.774} & \textit{0.261} & \textbf{0.306} \\
\cline{1-6}
\cline{1-6}
W-NOMINATE &  & 0.805 & 0.746 & \textbf{0.26} & \textit{0.309} \\
\cline{1-6}
\multirow[t]{2}{*}{PCA} & LR & 0.833 & 0.770 & 0.295 & 0.328 \\
 & MF & 0.797 & 0.745 & 0.29 & 0.321 \\
\cline{1-6}
TSNE & LR & 0.816 & 0.753 & 0.303 & 0.334 \\
\cline{1-6}
\multirow[t]{2}{*}{VAE} & 2-Layer & \textbf{0.845} & \textbf{0.774} & 0.291 & 0.322 \\
 & Logistic & \textit{0.844} & \textit{0.768} & 0.283 & 0.318 \\
\cline{1-6}
\bottomrule
\end{tabular}
\end{table}

\myparagraph{Train Fit} This scenario evaluates how well the existing answers can be recovered from the two-dimensional latent space for a given train sparsity $u$ and equivalent test sparsity. This setting is relevant where overfitting is not a concern, i.e., the goal is to represent existing data as accurately as possible. Examples in political science include the embedding of roll call data~\cite{boche_new_2018} and the dimensionality reduction of Polis~\cite{small2021polis}, where users are clustered in the latent space based on their binary answers to statements. We averaged the fit accuracies across all values of $u$ focusing on the binary data as used in the above-mentioned applications. We find that the VAE outperforms other models by reaching 84.5\% accuracy, regardless of whether the decision boundary of the decoder module is linear (Logistic) or non-linear (2-Layer). Specifically, the VAE achieves 4-5\% better results than W-NOMINATE or PCA which are currently used in these applications. 
\myparagraph{Train Impute} The second scenario focuses on missing value imputation in the train set. This is particularly important for the comment routing feature of Polis, where the objective is to select the most relevant next statement. For question selection, an accurate prediction of unanswered questions is required. Therefore, we computed the imputation accuracy of each spatial model for all sparse train sets with varying $u$. Our analysis reveals that
IDEAL and the non-linear VAE both achieved an accuracy of 77.4\%, surpassing the baseline (mean imputation) by around 11\%. Notably, this accuracy is significantly lower than their respective train-fit values, suggesting a tendency towards overfitting the given answers. This tendency is present for all tested models.  
\myparagraph{Test Fit} The next scenario examines each model's ability to generalize to test data. It naturally arises when a model trained on existing candidates must accommodate newcomers, e.g., when voters should be embedded in the same political space without modifying the dimensions obtained for the candidates alone (as required by the Smartmap of Smartvote). This setting was approached using the original Likert scale data. We used RMSE to assess how accurately the given answers of test users were recovered after embedding them with each spatial model. We find that IRT models exhibited the best generalization capability in this setting, with both W-NOMINATE and IDEAL outperforming other models with an RMSE of $0.26$.
\myparagraph{Test Impute} This last scenario is most relevant to the second part of our study. We assessed which spatial model best imputes missing values in the test set. Specifically, we trained the models using the complete train set with $u=0$ and measured the imputation RMSE for test sets across different values of $v$. This corresponds to the setting, when we want to predict the remaining answers of voters using a model trained on the answers of  candidates who completed the entire questionnaire. A comparison against the mean-imputation baseline highlighted the IDEAL model's superior performance, achieving an RMSE of 0.306 (compared to 0.380). However, all spatial models perform similarly well and lie within an interval of about 0.03. Despite this narrow margin the IDEAL model was chosen for the subsequent part of the experiment due to its consistent performance in all evaluated scenarios.

\subsection{Optimizing the Questionnaire}


In the second experiment, we aim to improve the order of the Smartvote questionnaire. For each selection method described in Section~\ref{sec:methods_selection}, we simulate the order in which questions would have been posed to voters. At each step, we compute the prediction accuracy and candidate recommendation accuracy (see Section~\ref{sec:methods_metrics}). As candidates, we use the fully complete train set ($u=0$), to which we fit the IDEAL model. As voters, we employ the corresponding test set, which we initialize as an empty matrix and then fill with each given answer until completion. We ran this experiment on a desktop computer, taking in total less than half an hour of wall time. For the analysis, we evaluate two different scenarios: individual questionnaires and matrix population.

\begin{figure}[t]
\centering
\includegraphics{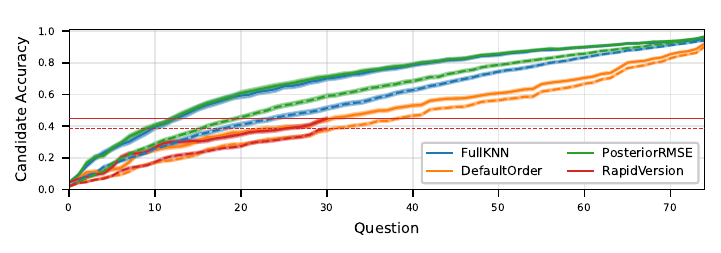}
\caption{Comparison of the candidate recommendation accuracy based on given answers only (dashed lines; Type I) and predicting remaining answers first (solid lines, Type II). Each line corresponds to the mean accuracy of all test users after the respective number of questions, where the uncertainty shading indicates the confidence interval of the mean estimator. The horizontal lines show the final accuracies of the \textsf{RapidVersion}. For clarity, only a selection of methods is shown.}
\label{fig:res_candidates}
\end{figure}

\myparagraph{Individual questionnaires}
In the first scenario, which resembles the typical framework of a VAA, each voter fills the questionnaire independently from others. After each answer, we compute the recommended candidates and compare them with the recommendation they would receive after having answered all questions (which we consider ground truth). In addition, we evaluate two different recommendation strategies. Type I only uses the given answers of a voter in each stage, reflecting the implementation as it is on Smartvote. Type II, our contribution, uses the IDEAL model to embed these given answers into the 2D space, where based on the location, the remaining answers are predicted. Then, this full range is used to compute the nearest neighbors in the candidate train set. 
Figure~\ref{fig:res_candidates} shows the average recommendation accuracy across all users (where the shaded area depicts the mean uncertainty). We can see, that the condensed questionnaire, here called \textsf{RapidVersion}, reaches around 40\% accuracy after selecting 31 questions in its fixed order (Type I recommendation). As a direct improvement of that, the same order of questions achieves 47\% percent recommendation accuracy when also using the predicted answers of the remaining questions (Type II). Similarly, \textsf{FullkNN}, \textsf{PosteriorRMSE}, and \textsf{DefaultOrder} achieve roughly 7-15\% more accurate recommendations when including the IDEAL predictions in the recommendation. This also holds for all other selection methods, which were omitted from the figure for clarity. 
Furthermore, it also becomes evident that after 9 questions, our \textsf{PosteriorRMSE} method achieves the same recommendation accuracy as the \textsf{RapidVersion} after 31 questions if using Type II recommendations. Likewise, after 31 questions, \textsf{PosteriorRMSE} would already have an accuracy of 74\%, which means that out of the 32 recommended candidates, around 24 also appear in the final recommendation (when the user answered all questions). However, in the \textsf{RapidVersion} recommendation, only 13 candidates are correct. This experiment therefore indicates that while all methods reach 100\% accuracy towards the end of the questionnaire, an adaptive method can drastically improve the recommendation accuracy in the case of early drop-out. Also, regardless of the ordering of the questionnaire, our Type II recommendation is on average more accurate than the current Type I recommendation.   

\myparagraph{Matrix population}

\begin{figure}[t]
\centering
\includegraphics{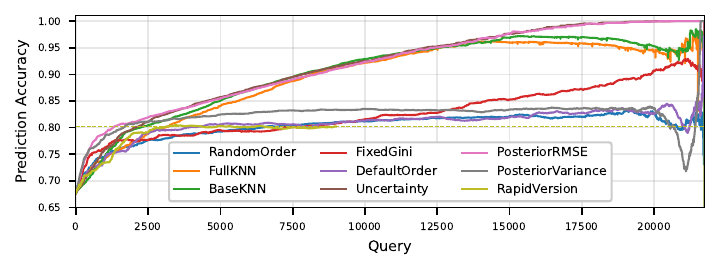}
\caption{Comparison of selection methods in an asynchronous setting. Each method can prioritize voter-question pairs where they assume acquired information most useful. In the calculation of the accuracy, only the remaining questions are considered.}
\label{fig:res_accuracy}
\end{figure}

In the second scenario, all users answer the questionnaire simultaneously. This situation is particularly relevant in a feature-value acquisition setting~\cite{saar-tsechansky_active_2009}, where the algorithm can determine which user-question pair is most valuable to obtain next. Here, we assess the methods' ability to prioritize users where predictions are more uncertain. After each selection of a user-question pair, we update the predictions and evaluate the accuracies. This process results in $290 \times 75 = 21,750$ queries, whereas the \textsf{RapidVersion} concludes after $290 \times 31 = 8,990$ queries due to its shorter questionnaire. In this scenario, we evaluate the accuracy of the predictions to the remaining unanswered questions only.
As Figure \ref{fig:res_accuracy} shows, all methods exceed the prediction accuracy of the \textsf{RapidVersion} after 8'990 queries. At that stage, four methods (\textsf{PosteriorRMSE}, \textsf{FullKNN}, \textsf{BaseKNN} and \textsf{Uncertainty}) distinguish themselves with a margin close to 10\% which later even increases. Notably, \textsf{PosteriorRMSE} is the first method to achieve the baseline of the \textsf{RapidVersion} (80\%) after only 1400 queries. The pattern towards the end of the figure is explained by the metric we chose: as we compare the accuracy of the remaining questions only (and in the end, there are very few unanswered questions), the variance can be quite large. \textsf{PosteriorRMSE} and \textsf{Uncertainty}, both designed to eliminate uncertainty early on, become certain about the remaining answers, explaining their early accuracy of close to 100\%. \textsf{PosteriorVariance}, which selects questions updating the position in the latent space, will ignore noisy questions and, thus,  get a noisy accuracy towards the end. These findings demonstrate the different objectives of the question selection methods. They suggest that Gini-based probabilistic methods are the most reliable in making question selections for adaptive questionnaires.

\begin{figure}[t]
\centering
\includegraphics{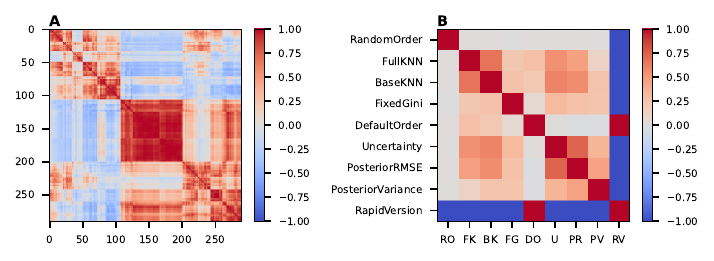}
\caption{Similarity of selection methods. \abc{A}~The order of the questionnaire is pairwise compared between the 290 test users. High values (red) indicate a strong correlation between rankings, whereas low values (blue) indicate anti-correlation. Users were sorted such that clusters become more visible. 
\abc{B}~The correlation between the order of questionnaires for each user is
pairwise compared for each selection method. The scores indicate the expected Spearman's Rank Correlation between two methods for a randomly selected user.}
\label{fig:res_correlation}
\end{figure}

\myparagraph{Adaptivity}
Finally, we assessed the adaptivity of each selection method by comparing the question order for each user. We did so in two ways: First, we counted the number of unique questions at each step. Second, we computed the Spearman's Rank Correlation (SRC) among all users. 
The first way of comparing the adaptivity of selection method, is to count the number of unique questions asked per iteration (assuming all users are asked simultaneously). The results (see Figure~\ref{fig:res_adaptivity} in the appendix) indicate that the adaptivity varies per selection method. Expectedly, the \textsf{RandomOrder} asks many unique questions per iteration (around 72), and the fixed-order methods ask one unique question per iteration. All other methods start with a unique question, and based on the answer of each user, increase the number of questions per iteration, reaching a maximum spread of 50-70 questions. Therefore, even for adaptive methods, there is an initial period that is less adaptive. Towards the end of the questionnaire, these methods narrow down their spread to a fewer number of questions, which likely correspond to those that yield the least information. Among these methods, we find that the \textsf{BaseKNN} and \textsf{FullKNN} are the most adaptive, followed by \textsf{PosteriorRMSE}, \textsf{PosteriorVariance}, and \textsf{Uncertainty}.
This analysis, however, only compares methods at each step. A more thorough analysis is shown in Figure \ref{fig:res_correlation}A. Here, given a specific method, we compute the SRC among all users. This score measures the correlation of the order of the questions for pairs of users. We see that there exist different clusters of users with similar questionnaires. These clusters likely correspond to users' political orientation. Computing these matrices for all selection methods and then averaging them, we get scores of the adaptivity, which confirm the previous results and are shown in the appendix in Figure~\ref{fig:adap_bar}.  
Finally, we compute the similarity of questionnaire orderings between the selection methods. For each pair of methods, we compute the average SRC for all users, and average them to a similarity score. Figure \ref{fig:res_correlation}B shows that the \textsf{RapidVersion}, expectedly, anti-correlates with all methods, given its shorter length. \textsf{RandomOrder} shows no correlation with any of the other methods. More interestingly, the \textsf{BaseKNN} and \textsf{FullKNN} show a higher correlation to each other than to the rest of the selection methods, while \textsf{PosteriorRMSE} and \textsf{Uncertainty} also correlate with each other. This shows, that methods with similar concepts also yield similar questionnaire orderings. 

\section{Conclusions}

In this work, we compared seven different spatial models in the context of political survey data such as VAAs or wiki surveys. In the first part, we evaluated how well these spatial models can impute missing values. In the second part, we used the best-performing algorithm (IDEAL) to introduce active learning to VAAs, enhancing the questionnaire to acquire information more effectively.

In the first part of the study, we discovered that the VAE outperforms other algorithms in accurately embedding the train set. Consequently, we propose replacing the use of PCA with VAEs in both Smartvote and Polis, as VAEs provide a superior data representation. This advantage is not only in terms of more accurate decision boundaries but also in the more precise prediction of answers for users with missing values. Additionally, we find that IRT models perform well with Likert scale data, despite being originally designed for binary settings. The IDEAL algorithm imputed missing values with the lowest RMSE among all evaluated algorithms, leading us to select this model for the second part of our study. Overall, we conclude that our proposed methodology for evaluating these spatial models can serve as a benchmark for testing other dimensionality reduction approaches in the context of survey data.

In the second part, we evaluated the effectiveness of different methods to select the next question in the Smartvote questionnaire. We compared how fast the candidate recommendation accuracy of an adaptive questionnaire reaches the baseline of asking 31 manually selected questions (contained in the condensed version of today's Smartvote questionnaire). We find that after 9 questions, we achieve the same accuracy as this \textsf{RapidVersion}. We also find that we reach 74\% accuracy after 31 asked questions. Here the \textsf{PosteriorRMSE} method yielded the best results by recommending almost twice as many correct candidates. 
In addition, we show that VAAs should use a different candidate recommendation procedure: instead of suggesting the nearest neighbor candidates based on the currently recorded answers only, these should be used to predict the answers to the not-yet considered questions and use the resulting ``full'' set of answers to find the nearest candidates for consideration.  
This consistently improves the recommendation accuracy for all selection methods, including fixed orders.

In summary, our work demonstrates a powerful application of machine learning methods in the context of VAAs---a societally highly relevant domain---and can improve their efficacy in many different ways. 
Combined with the spatial model used to predict unanswered questions, this is a novel enhancement of survey methodology, which showed drastic improvements in recommendation accuracy. Especially in the binary context (such as used by Polis), where arguably a fixed-order method is used for comment routing, our algorithm is directly applicable and could improve the underlying analysis, e.g., clustering results. 
Future work should extend the selection methods to Likert scale data, benchmarking these algorithms with other VAA datasets or validating adaptive questionnaires in real-world elections. Here, the focus should not only be candidate recommendation accuracy but also voters' trust in such adaptive questionnaires. We believe that such research will make VAAs more effective, and, therefore, contribute to political salience in the realm of digital democracy. 

\ifthenelse{\boolean{anonymize}}
    {
    }{
\subsubsection{Acknowledgements} We would like to thank the team from Politools for providing the Smartvote data. Also, we acknowledge the Swiss National Science Foundation (SNSF), which funds the research with grant ID CRSII5-205975. 
    }
%
%
%
\bibliographystyle{splncs04}
\bibliography{bibliography/AdjustedLibrary}

\begin{thebibliography}{10}
\providecommand{\url}[1]{\texttt{#1}}
\providecommand{\urlprefix}{URL }
\providecommand{\doi}[1]{https://doi.org/#1}

\bibitem{bachmann_wasserstein_2023}
Bachmann, F., Hennig, P., Kobak, D.: Wasserstein t-{SNE}. In: Machine {Learning} and {Knowledge} {Discovery} in {Databases}, vol. 13713, pp. 104--120. Springer International Publishing, Cham (2023)

\bibitem{boche_new_2018}
Boche, A., Lewis, J.B., Rudkin, A., Sonnet, L.: The new {Voteview}.com: preserving and continuing {Keith} {Poole}’s infrastructure for scholars, students and observers of {Congress}. Public Choice  \textbf{176}(1-2),  17--32 (Jul 2018)

\bibitem{clinton_statistical_2004}
Clinton, J., Jackman, S., Rivers, D.: The {Statistical} {Analysis} of {Roll} {Call} {Data}. American Political Science Review  \textbf{98}(2),  355--370 (May 2004)

\bibitem{garzia_voting_2019}
Garzia, D., Marschall, S.: Voting {Advice} {Applications}. In: Oxford {Research} {Encyclopedia} of {Politics}. Oxford University Press (Mar 2019)

\bibitem{garzia_voting_2017}
Garzia, D., Trechsel, A.H., De~Angelis, A.: Voting {Advice} {Applications} and {Electoral} {Participation}: {A} {Multi}-{Method} {Study}. Political Communication  \textbf{34}(3),  424--443 (Jul 2017)

\bibitem{jackman2024pscl}
Jackman, S.: {pscl}: Classes and Methods for {R} Developed in the Political Science Computational Laboratory. University of Sydney, Sydney, Australia (2024)

\bibitem{kingma_auto-encoding_2022}
Kingma, D.P., Welling, M.: Auto-{Encoding} {Variational} {Bayes} (Dec 2022)

\bibitem{van2008visualizing}
Van~der Maaten, L., Hinton, G.: Visualizing data using {t-SNE}. Journal of machine learning research  \textbf{9}(11) (2008)

\bibitem{mccoy_variational_2018}
McCoy, J.T., Kroon, S., Auret, L.: Variational {Autoencoders} for {Missing} {Data} {Imputation} with {Application} to a {Simulated} {Milling} {Circuit}. IFAC-PapersOnLine  \textbf{51}(21),  141--146 (2018)

\bibitem{montgomery_computerized_2013}
Montgomery, J.M., Cutler, J.: Computerized {Adaptive} {Testing} for {Public} {Opinion} {Surveys}. Political Analysis  \textbf{21}(2),  172--192 (2013)

\bibitem{orsini2022machine}
Orsini-Rosenberg, D.: Machine {Learning} vs. {Swiss} {Politics}. Bachelor's Thesis  (2022)

\bibitem{pianzola_impact_2019}
Pianzola, J., Trechsel, A.H., Vassil, K., Schwerdt, G., Alvarez, R.M.: The {Impact} of {Personalized} {Information} on {Vote} {Intention}: {Evidence} from a {Randomized} {Field} {Experiment}. The Journal of Politics  \textbf{81}(3),  833--847 (Jul 2019)

\bibitem{smartvote2019data}
{Politools}: {D}aten zu den {N}ationalrats- und {S}tänderatswahlen 2019 der {O}nline-{W}ahlhilfe smartvote.ch (2019)

\bibitem{policar_opentsne_2019}
Poličar, P.G., Stražar, M., Zupan, B.: {openTSNE}: a modular {Python} library for t-{SNE} dimensionality reduction and embedding. preprint, Bioinformatics (Aug 2019)

\bibitem{poole2007scaling}
Poole, K., Lewis, J., Lo, J., Carroll, R.: Scaling {Roll} {Call} {Votes} with wnominate in {R}. Journal of Statistical Software  (2007)

\bibitem{poole_spatial_1985}
Poole, K.T., Rosenthal, H.: A {Spatial} {Model} for {Legislative} {Roll} {Call} {Analysis}. American Journal of Political Science  \textbf{29}(2), ~357 (May 1985)

\bibitem{potthoff_estimating_2018}
Potthoff, R.: Estimating {Ideal} {Points} from {Roll}-{Call} {Data}: {Explore} {Principal} {Components} {Analysis}, {Especially} for {More} {Than} {One} {Dimension}? Social Sciences  \textbf{7}(2), ~12 (Jan 2018)

\bibitem{reckase2006irt}
Reckase, M.D.: Multidimensional item response theory. In: Psychometrics, Handbook of Statistics, vol.~26, pp. 607--642. Elsevier (2006)

\bibitem{saar-tsechansky_active_2009}
Saar-Tsechansky, M., Melville, P., Provost, F.: Active {Feature}-{Value} {Acquisition}. Management Science  \textbf{55}(4),  664--684 (Apr 2009)

\bibitem{sigfrid_irt_2024}
Sigfrid, K.: {IRT} for voting advice applications: a multi-dimensional test that is adaptive and interpretable. Quality \& Quantity  (Mar 2024)

\bibitem{small2021polis}
Small, C., Bjorkegren, M., Erkkil{\"a}, T., Shaw, L., Megill, C.: Polis: Scaling deliberation by mapping high dimensional opinion spaces. Recerca: revista de pensament i an{\`a}lisi  \textbf{26}(2) (2021)

\bibitem{zou_selective_2018}
Zou, H., Xue, L.: A {Selective} {Overview} of {Sparse} {Principal} {Component} {Analysis}. Proceedings of the IEEE  \textbf{106}(8),  1311--1320 (Aug 2018)

\end{thebibliography}
%
\newpage
\appendix
\section{General}
\label{app:overview}

\begin{figure}[h]
    \centering
    \begin{tikzpicture}
        \node[anchor=north west, inner sep=0] (image) at (1,0) {\includegraphics[height=0.4\textwidth]{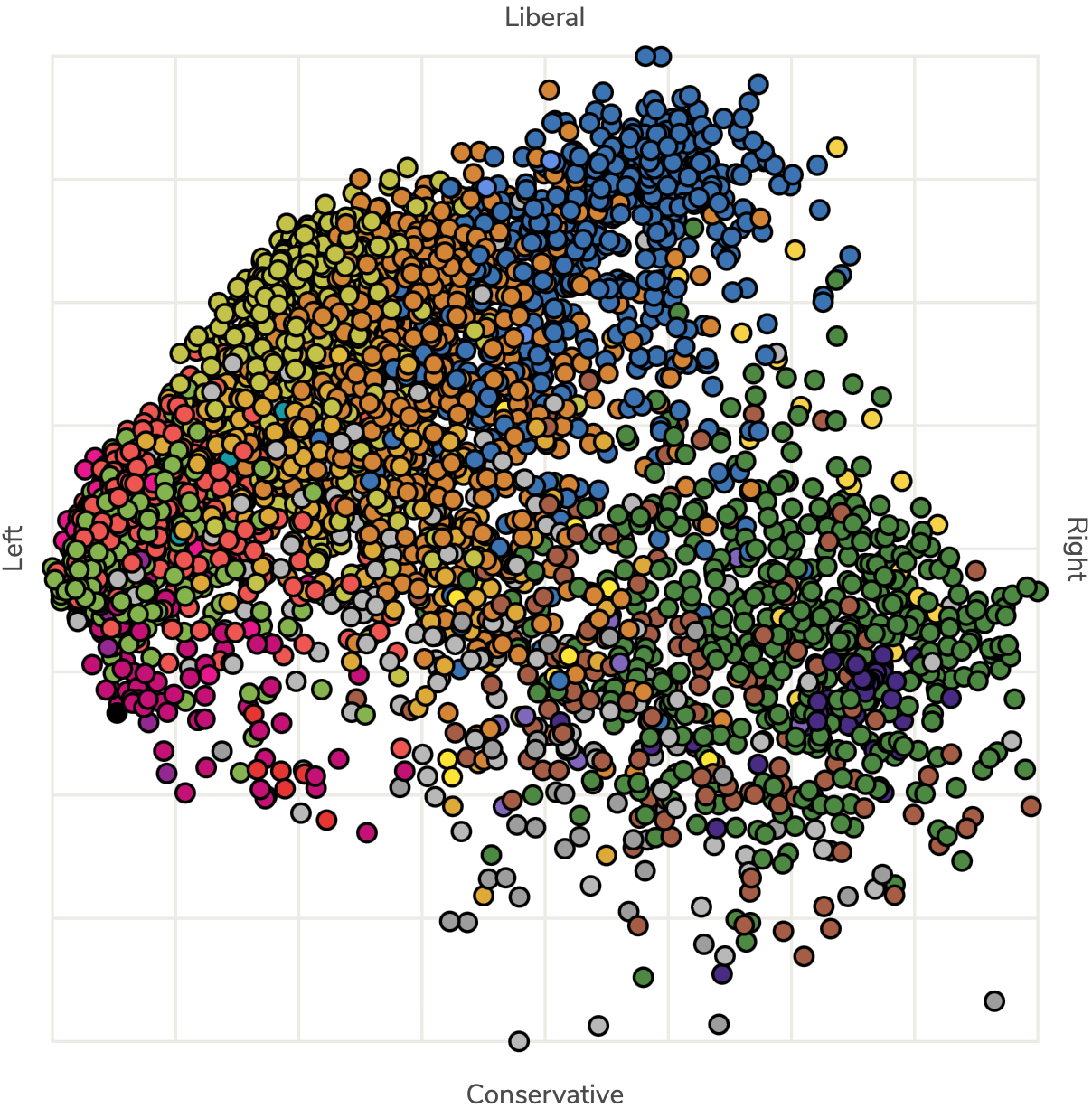}};
        \node[anchor=north east, font=\sffamily\bfseries\small] at (image.north west) {A};
        \node[anchor=north east, inner sep=0] (imageB) at (\textwidth,0) {\includegraphics[height=0.4\textwidth]{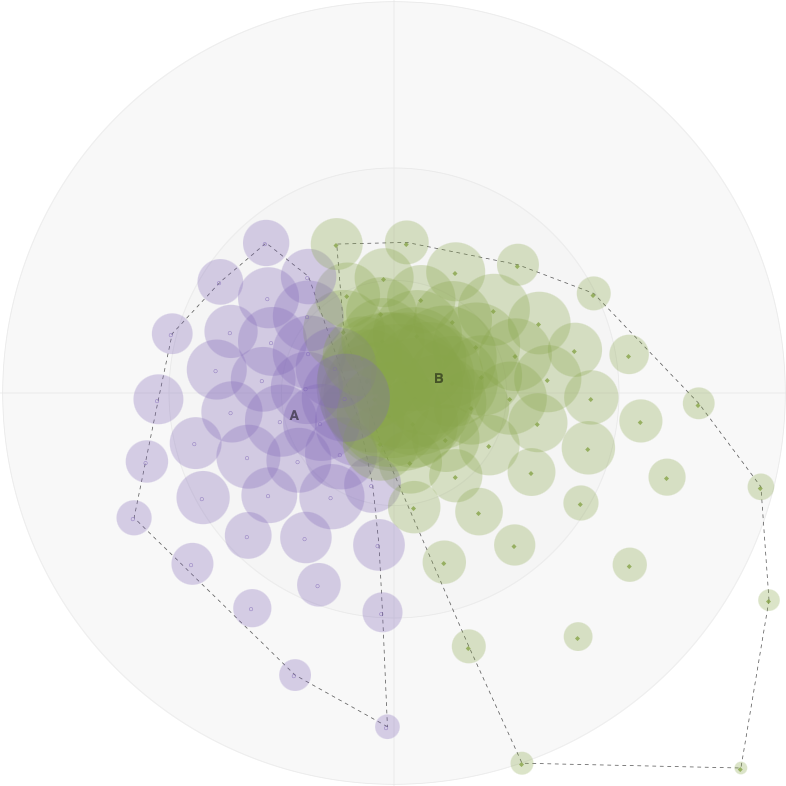}};
        \node[anchor=north east, font=\sffamily\bfseries\small] at (imageB.north west) {B};
    \end{tikzpicture}
    \caption{Spatial models in political survey applications. \abc{A}~The VAA Smartvote visualizes the political landscape as a two-dimensional PCA embedding of candidates' answers to their survey. The dots represent candidates and their respective party, e.g. dark green for SVP and red for SP. Image taken from \mbox{\url{https://smartvote.ch}}. 
    \abc{B}~In the wiki survey Polis, users like or dislike statements that are shown to them sequentially. Based on their interaction, they are projected into a two-dimensional PCA-space. Clusters (gray lines) are identified with k-means. Image taken from \url{https://pol.is}.}
    \label{fig:applications}
\end{figure}

\begin{figure}[h!]
    \centering
    \includegraphics[width=\textwidth]{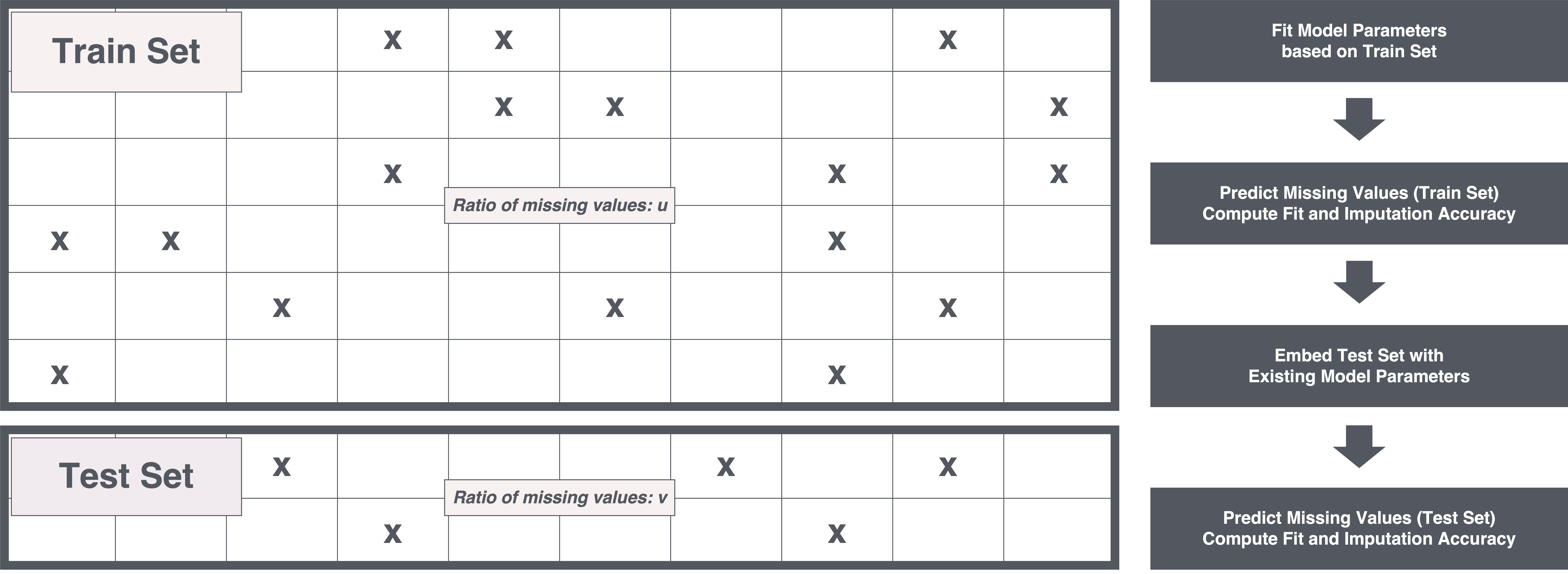}
    \caption{Experiment Design. The complete Smartvote candidates dataset is split into a train and test set, where a fraction $u$ (and $v$ respectively) of values are removed. For different $u$, each spatial model is then fitted to the sparse train set. Using the trained models' parameters, the test set is then embedded into the same latent space. Based on the embedding of each candidate, all answers are then predicted. Fit scores correspond to how well the given data matches the predicted data, whereas impute scores correspond to the correct prediction of missing data.}
    \label{fig:schema}
\end{figure}

\newpage
\section{Alternative Smartmaps}
\label{app:models}

\begin{figure}[h]
\centering
\includegraphics{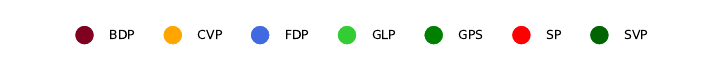}
\caption{Legend for all the embeddings of the train and test sets. The colors correspond to the seven major Swiss parties.}
\label{fig:legend}
\end{figure}

\begin{figure}[h!]
\centering
\includegraphics{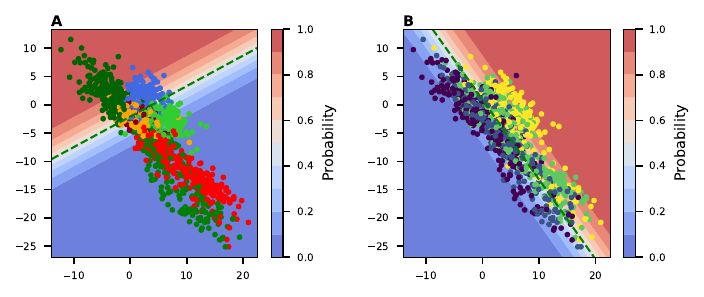}
\caption{Latent space of the VAE with a logistic decoder. The train-set candidates embedding of the original Smartvote dataset is overlaid with predictions for two questions. \abc{A}~Question 46: \enquote{Should high traffic motorways be expanded to six lanes?} Here, the colors represent party affiliation.  \abc{B}~Question 28: \enquote{Should the rules for reproductive medicine be further relaxed?} Here, the colors represent candidates' answers on a 4-Likert scale from \enquote{Fully Agree} (yellow) to \enquote{Fully Disagree} (dark blue).}
\label{fig:vae_log}
\end{figure}

\begin{figure}[p]
\centering
\includegraphics{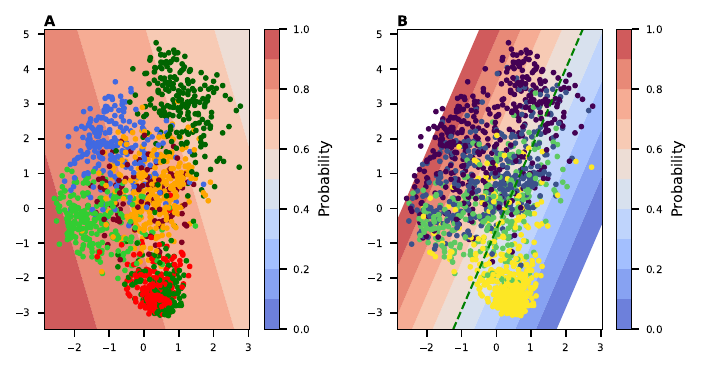}
\caption{PCA projection of the original Smartvote dataset overlaid  with MF predictions. \abc{A}~Question 70: \enquote{Should the federal government spend more or less in the area of Education and research?} Here, the colors represent party affiliation.  \abc{B}~Question 12: \enquote{Should the federal government provide more support for the construction of non-profit housing?} Here, the colors represent candidates' answers on a 4-Likert scale from \enquote{Fully Agree} (yellow) to \enquote{Fully Disagree} (dark blue). The MF predictions (background color) were capped at 0 and 1.}
\label{fig:pca_mf}
\end{figure}

\begin{figure}[hp]
\centering
\includegraphics{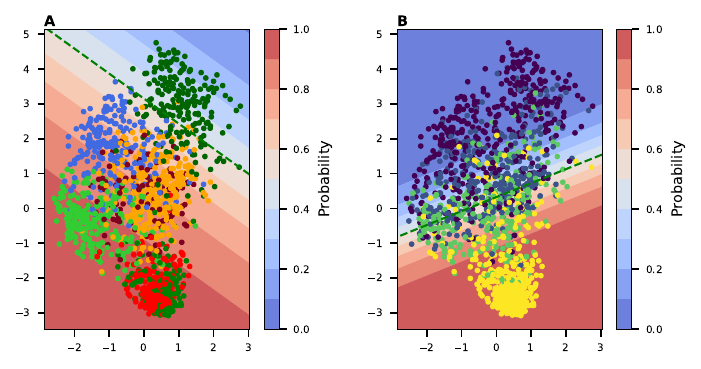}
\caption{PCA projection of the original Smartvote dataset overlaid  with LR predictions. \abc{A}~Question 70: \enquote{Should the federal government spend more or less in the area of Education and research?} Here, the colors represent party affiliation.  \abc{B}~Question 12: \enquote{Should the federal government provide more support for the construction of non-profit housing?} Here, the colors represent candidates' answers on a 4-Likert scale from \enquote{Fully Agree} (yellow) to \enquote{Fully Disagree} (dark blue). The LR model clearly captures the structure of candidates' answers better than the MF.}
\label{fig:pca_lr}
\end{figure}

\begin{figure}[t]
\centering
\includegraphics{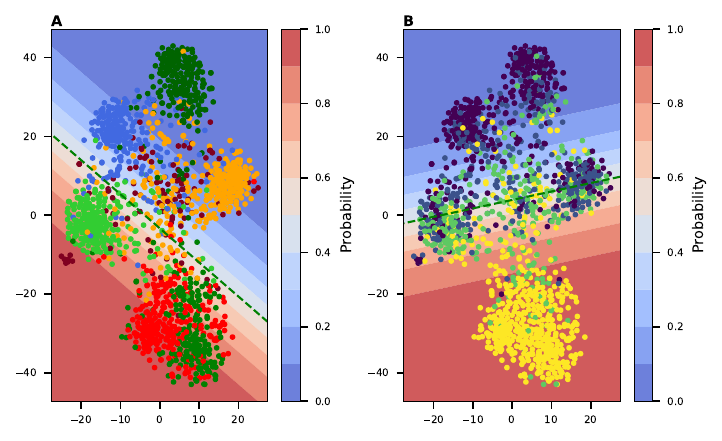}
\caption{$t$-SNE projection of the original Smartvote dataset overlaid with LR predictions. \abc{A}~Question 55: \enquote{Are you in favour of lowering the voting age to 16?} Here, the colors represent party affiliation.  \abc{B}~Question 12: \enquote{Should the federal government provide more support for the construction of non-profit housing?} Here, the colors represent candidates' answers on a 4-Likert scale from \enquote{Fully Agree} (yellow) to \enquote{Fully Disagree} (dark blue).}
\label{fig:tsne_lr}
\end{figure}

\clearpage

\section{Mathematical framework}
\label{app:selection}

In the section we provide more details on the mathematical framework used in both our experiments. In particular, we will explain the selection methods from Section~\ref{sec:methods_selection} based on the IDEAL model described in Section~\ref{sec:methods_spatial}.

\subsection{Gini impurity}
First, we derive the Gini impurity for binary probabilities. By definition, Gini impurity is
\begin{equation}
G(\vec{p}) = 1 - \sum_{i=1}^{2} p_i^2
\end{equation}
for a vector of probabilities $\vec{p}$ which sums to 1. In the binary case, it therefore holds that
\begin{align}
G(p) &=  1 - (p^2 + (1-p)^2) \\
     &= 1 - (p^2 + 1 - 2p + p^2) \\
     &= 1 - 2p^2 + 2p - 1 \\
     &= 2p - 2p^2 \\
     &= 2p(1 - p).
\end{align}
\subsection{IDEAL}
For the probabilistic question selection methods, we need to define the prior probabilities and likelihood of the spatial model. We assume a normal prior
\begin{equation}
    P(X) \sim \mathcal{N}(X; \mu, \Sigma),
\end{equation}
and write the likelihood function for a single question $Y_k$
\begin{equation}
    P(Y_k=y_k|X) = \sigma_k(X)^{y_k} \cdot (1-\sigma_k(X))^{1-y_k},
\end{equation}
according to the IDEAL model described in Equation~\ref{eq:ideal} in Section~\ref{sec:methods_spatial}
\begin{equation}
    \sigma_k(X) = \Phi(X^T \beta_k - \alpha_k),
\end{equation}
where in our two dimensional case $X^T \beta_k$ is the dot product of the model discrimination parameters $\beta$ and the spatial location $X$. 
We can then compute the evidence by integrating over $X$ in the joint distribution 
\begin{equation}
   P(Y) = \int P(Y|X)\cdot P(X) \, \mathrm{d}X
\end{equation}
which yields the marginal probabilities in $Y$.
Note, that the IDEAL model implies conditional independence of questions given the spatial location $X$, so it holds
\begin{equation}
    P(Y_{I\cup N}|X) = P(Y_I|X) \cdot P(Y_N|X).
\end{equation}
The model predictions for unanswered questions $Y_N$ given some initial answers $Y_I$ then become
\begin{align}
    P(Y_N|Y_I) &= \int P(X,Y_N|Y_I) \, \mathrm{d}X\\
    &= \int P(Y_N|X,Y_I)\cdot P(X|Y_I) \, \mathrm{d}X\\
    &= \int P(Y_N|X)\cdot P(X|Y_I) \, \mathrm{d}X,
\end{align}
using in the law of total probability in the first line and conditional independence in the last line. With Bayesian inference we can then compute the posterior distribution of $X$ given some set of given answers $Y_I$
\begin{equation}
    P(X|Y_I) =  \frac{P(Y_{I}|X) \cdot P(X)}{P(Y_I)}.
\end{equation}
This can be computed iteratively as we can see that adding an additional answer $Y_k$ merely means multiplying by another factor 
\begin{align}
    P(X|Y_{I\cup k}) &=  \frac{P(Y_{I\cup k}|X) \cdot P(X)}{P(Y_{I\cup k})}\\
    &= \frac{P(Y_I|X) \cdot P(Y_k|X) \cdot P(X)}{P(Y_k|Y_I) \cdot P(Y_I)}\\
    &=P(X|Y_I) \frac{P(Y_k|X)}{P(Y_k|Y_I)}.
\end{align}
This is relevant for question selection, as the simulation of posteriors under possible answers can be efficiently computed. It is important to note, that in our analysis we discretize the space and restrict coordinates to a $[-3, 3]^2$ interval. Therefore, all computations can be done with vector operations in \textsf{numpy} which is efficient for the otherwise intractable normalization constants. 

\clearpage

\section{Detailed results}
\label{app:results}

\begin{figure}[h]
\centering
\includegraphics{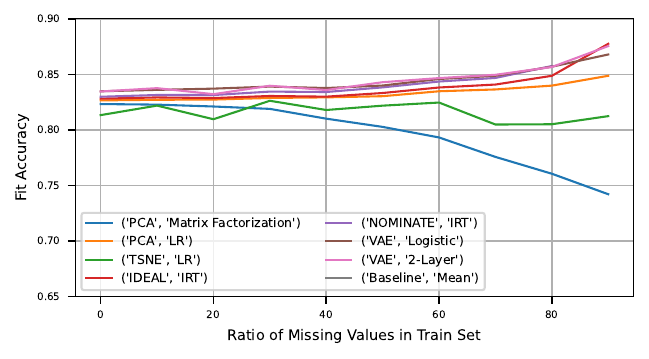}
\caption{Accuracy results in the \textit{Train Fit} scenario. All models reconstruct values similarly well for low train sparsity. For higher sparsity, PCA and $t$-SNE fail to reconstruct the data well when compared to the results of IRT and VAE.}
\label{fig:trainfit}
\end{figure}

\begin{figure}[h!]
\centering
\includegraphics{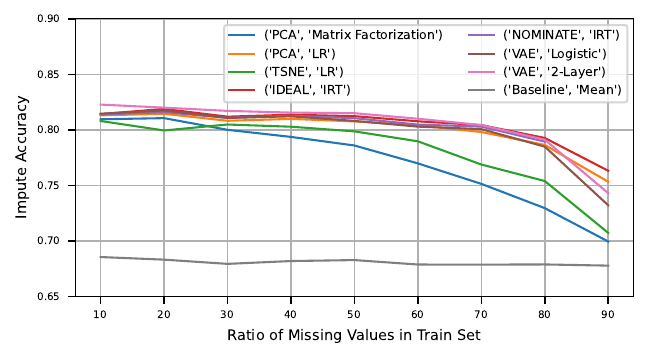}
\caption{Accuracy results in the \textit{Train Impute} scenario. All models impute missing values similarly well for low train sparsity. For higher sparsity, all models become less accurate, however, PCA and $t$-SNE being especially bad, almost reaching the mean-imputation baseline.}
\label{fig:trainimpute}
\end{figure}

\begin{figure}[hp]
\centering
\includegraphics{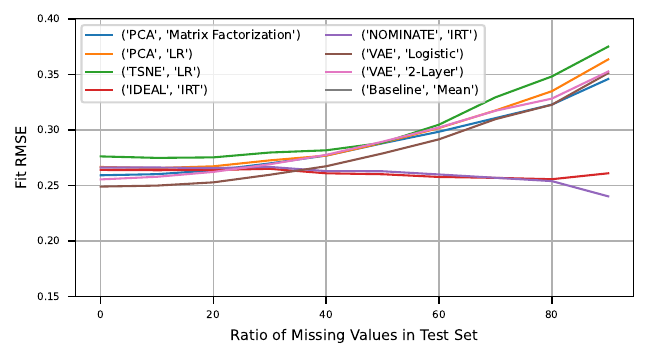}
\caption{RMSE results in the \textit{Test Fit} scenario. All models were trained on the complete train set. They reconstruct the data similarly well for low test sparsity. For higher sparsity, the IRT models distinguish themselves from other models with a significantly lower RMSE.}
\label{fig:testfit}
\end{figure}

\begin{figure}[hp]
\centering
\includegraphics{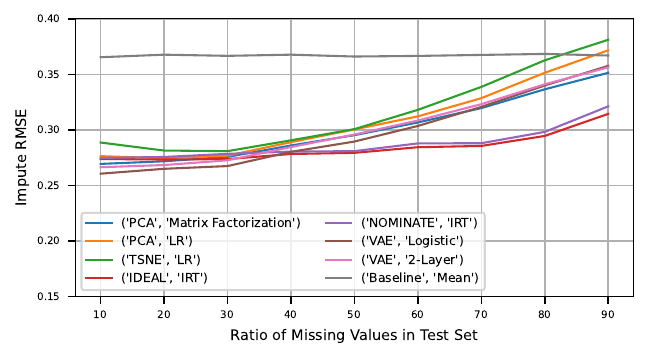}
\caption{Accuracy results in the \textit{Test Impute} scenario. All models were trained on the complete train set. They impute missing values in the test data similarly well for low test sparsity. For higher sparsity, the IRT models distinguish themselves from other models with a significantly lower RMSE.}
\label{fig:testimpute}
\end{figure}

\begin{figure}[t]
\centering
\includegraphics{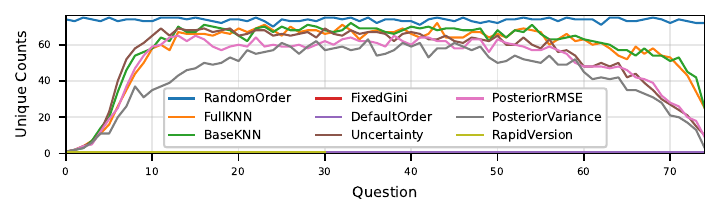}
\caption{Adaptivity of different selection methods. The number of unique questions across all users per step is compared for different methods. The random order shows a high adaptivity, whereas fixed order methods show one question per step.}
\label{fig:res_adaptivity}
\end{figure}

\begin{figure}[hp]
\centering
\includegraphics{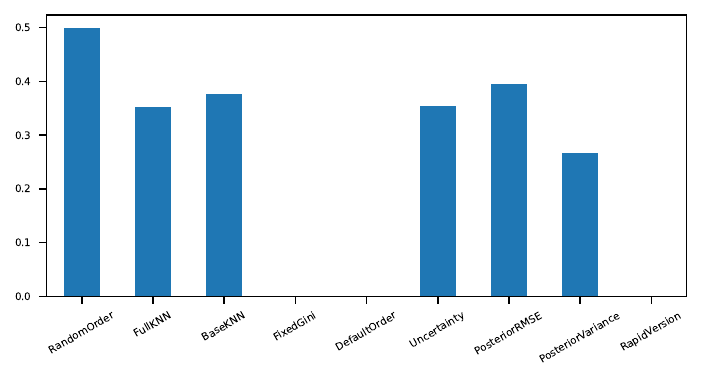}
\caption{Adaptivity scores of different selection methods. The scores incorporates  the average SRC among all users. This quantifies the expected correlation of the question ranking  for pairs of users. Fixed-order method have no adaptivity, while the \textsf{RandomOrder} is the most adaptive. The other methods have similar scores.}
\label{fig:adap_bar}
\end{figure}

\end{document}